\documentclass[10pt,twocolumn,letterpaper]{article}

\usepackage[pagenumbers]{cvpr} 
\usepackage{microtype}
\usepackage{multirow}
\usepackage{marvosym}
\definecolor{cvprblue}{rgb}{0.21,0.49,0.74}
\usepackage[pagebackref,breaklinks,colorlinks,allcolors=cvprblue]{hyperref}

\def\paperID{4355} 
\def\confName{CVPR}
\def\confYear{2025}

\newcommand{\mymethod}[1]{\textbf{OOD-HOI}}
\title{OOD-HOI: Text-Driven 3D Whole-Body Human-Object Interactions Generation Beyond Training Domains}

\author{Yixuan Zhang$^{1}$, Hui Yang$^{1}$, Chuanchen Luo$^{4}$, Junran Peng$^{5}$, Yuxi Wang$^{1 \textsuperscript{\Letter}}$, Zhaoxiang Zhang$^{1,2,3}$ \\
{\normalsize\centering$^{1}$ Centre for Artificial Intelligence and Robotics, HKISI, CAS}
{\normalsize\centering$^{2}$ Institute of Automation, Chinese Academy of Sciences}\\
{\normalsize\centering$^{3}$ University of Chinese Academy of Sciences} 
{\normalsize\centering$^{4}$ Shandong University} 
{\normalsize\centering$^{5}$ University of Science and Technology Beijing}}

\begin{document}


\twocolumn[{%
\renewcommand\twocolumn[1][]{#1}%
\maketitle
\vspace{-1.8em}
\includegraphics[width=.98\linewidth,trim=10 90 30 40, clip]{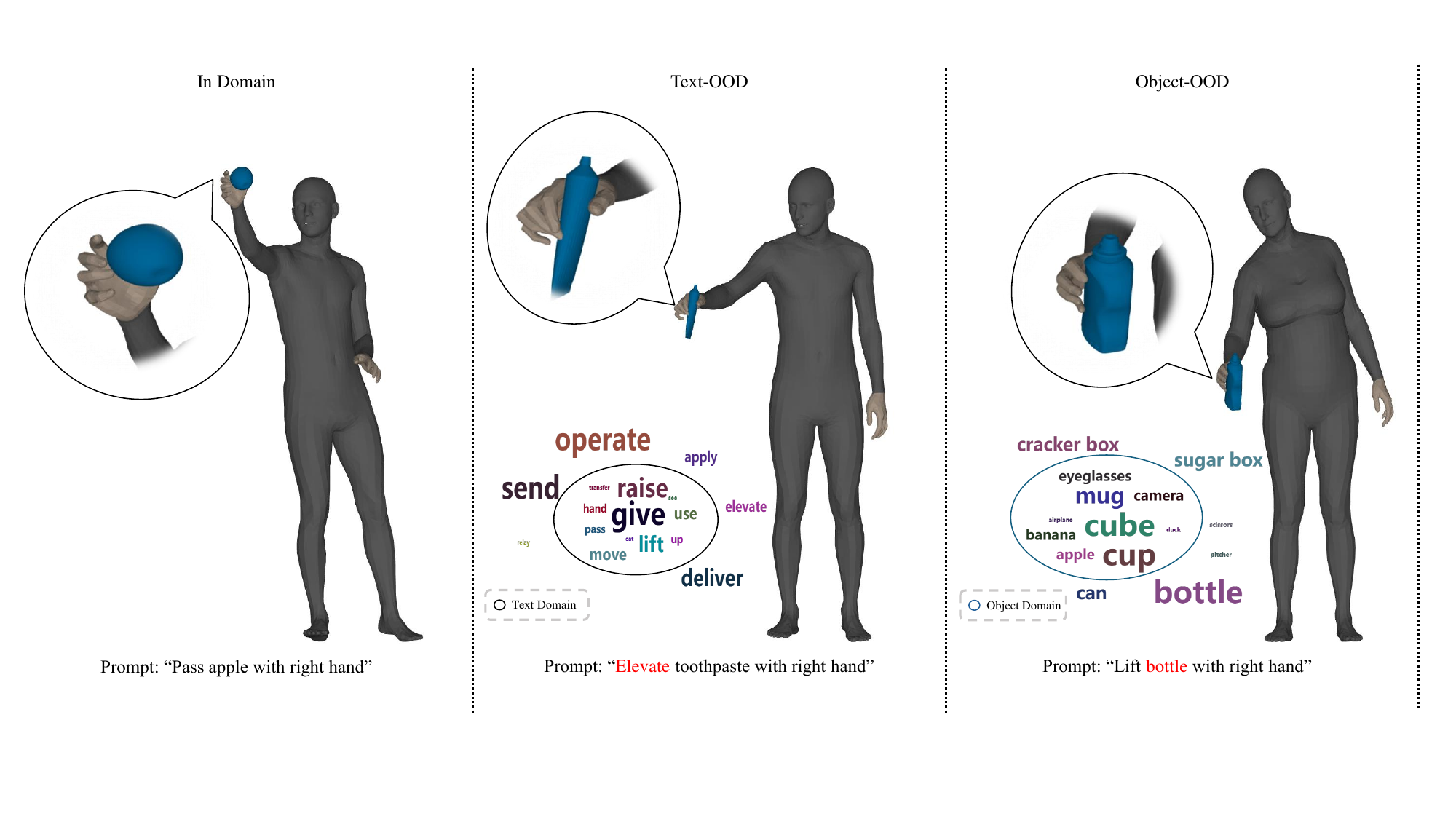}
\vspace{-1em}
\captionof{figure}{We introduce OOD-HOI, a novel text-driven method for generating 3D human-object interactions in out-of-domain scenarios. OOD-HOI can generate realistic whole body human-object interaction poses directly from textual descriptions. Even when encountering unseen entities in instructions (highlighted in red), such as the action \textit{elevate} or the object \textit{bottle}, it produces physically plausible results.}
\label{fig:teaser}
\vspace{2em}
}]

\begin{abstract}
Generating realistic 3D human-object interactions (HOIs) from text descriptions is a active research topic with potential applications in virtual and augmented reality, robotics, and animation. However, creating high-quality 3D HOIs remains challenging due to the lack of large-scale interaction data and the difficulty of ensuring physical plausibility, especially in out-of-domain (OOD) scenarios. Current methods tend to focus either on the body or the hands, which limits their ability to produce cohesive and realistic interactions. In this paper, we propose \mymethod{OOD-HOI}, a text-driven framework for generating whole-body human-object interactions that generalize well to new objects and actions. Our approach integrates a dual-branch reciprocal diffusion model to synthesize initial interaction poses, a contact-guided interaction refiner to improve physical accuracy based on predicted contact areas, and a dynamic adaptation mechanism which includes semantic adjustment and geometry deformation to improve robustness. Experimental results demonstrate that our \mymethod{OOD-HOI} could generate more realistic and physically plausible 3D interaction pose in OOD scenarios compared to existing methods.  The project page along
with videos can be found at:\href{https://nickk0212.github.io/ood-hoi/}{https://nickk0212.github.io/ood-hoi/}
\end{abstract}    
\section{Introduction}

Text-driven synthesis of 3D human-object interactions (HOIs) has attracted much attention in recent years, aiming to generate coherent and semantically meaningful interactions between humans and objects \cite{lv2025himo, zhang2025nl2contact, jian2023affordpose}. 
These accurate 3D HOI methods show significant application potential across various fields, such as virtual reality, augmented reality, animation, robotics, and embodied AI. Therefore, it is essential to create physically plausible and contextually relevant 3D scenes for a variety of applications.  

Despite the notable efforts, 3D human-object interaction generation from textual descriptions still faces significant challenges due to the scarcity of large-scale interaction data and physical prior. Concretely, compared to 3D human motion generation, existing human-object interaction datasets~\cite{christen2024diffh2o,bhatnagar22behave} are approximately 10$\times$ smaller~\cite{mahmood2019amass,Guo_2022_CVPR}. Additionally, these datasets are collected from specific scenarios (\textit{e.g., motion capture or simulations}). Recent advanced text-to-3D HOI methods~\cite{christen2024diffh2o} have shown significant performance degradation in out-of-domain scenarios, particularly in terms of diversity and controllability when guided by fine-grained textual descriptions or image-based pose estimators~\cite{fan2023arctic}. On the other hand, generating geometrically and semantically plausible full-body avatars grasping objects is challenging, owing to the inherent ambiguity of text description. For example, grasping the mug from the top or the side is valid by prompting a motion of \textit{``pick up the mug''}. 

To address above problems, existing text-driven 3D HOI synthesis methods involve extending existing diffusion models \cite{ramesh2022hierarchical, tevet2023human, ho2020denoising}, to generate human-object interactions, 
which can be roughly categorized into two streams: body-centric and hand-centric approaches. The former approaches~\cite{xu2023interdiff, peng2023hoi, li2023object} focus on modeling body-object contact, capturing the general motion trend of interactions but often ignoring the precise role of the hands. This omission limits their ability to generate fully realistic interactions, as hands play a crucial role in manipulating objects. On the other hand, hand-centric methods synthesize hand-grasping motions~\cite{cha2024text2hoi, christen2024diffh2o} but disregard the rest of the body, leading to unrealistic results where hand movements are not aligned with the body’s posture and context of the interaction. 

Another important issue in 3D human-object interaction generation lies in handling out-of-domain (OOD) generalization. It arises when the trained model encounters prompts that diverge significantly from its training data, such as novel objects, unusual actions, or unique human-object combinations that the model has not seen before. GraspTTA~\cite{jiang2021hand} proposed a test-time adaptation paradigms for generalize in out-of-domain objects, and InterDreamer~\cite{xu2024interdreamer} generate corresponding plausible human-object interactions by the help of large language model. Although these methods have achieved generalization ability at the usual action level, they still struggle to produce reasonable and contextual interactions at the novel object level.

In this paper, we provide a novel text-driven 3D human-object interaction method for out-of-domain generation, named \mymethod{OOD-HOI}, by jointly considering the generation of the human body, hands, and objects. Specifically, the proposed \mymethod{OOD-HOI} includes three components: a dual-branch reciprocal diffusion model, a contact-guided interaction refiner, and a dynamic adaptation. Specifically, the dual-branch reciprocal diffusion module aims to generate whole body interactions in a compositional way from a text description and object point clouds. The contact-guided interaction refiner revises the interaction pose by integrating the predicted contact area as guidance that can be applied at inference time to the diffusion process. To improve the generalization when encountering our method for out-of-domain HOI generation, we propose a dynamic adaptation method to enhance the generation of unseen objects and various text descriptions. Specifically, we argue that the generalization focuses on the alignment between text description and actions, as well as various objects. Therefore, we introduce a semantic adjustment and geometry deformation to tackle Text-OOD and Object-OOD problems, respectively.

Our contributions are summarized as follows:
\begin{enumerate}
    \item We propose \mymethod{OOD-HOI}, a novel text-driven whole-body human-object interactions generation method by joint considering the information exchange from the human body, hands and objects.
    \item Our proposed contact-guided interaction refiner improves interaction poses by integrating predicted contact areas as guidance. This refiner enhances pose refinement by applying physical plausible contact-guided updates during the diffusion process, thus enabling more realistic and contextually accurate interactions.
    \item To enhance generalization across unseen objects and diverse text prompts, we introduce a dynamic adaptation mechanism, incorporating semantic adjustment and geometry deformation. This approach includes the strategic replacement of action prompts with synonymous terms, alongside controlled deformation of object geometries under contact constraints.
\end{enumerate}
\section{Related Works}

\subsection{Human Motion Generation}
Research on human motion generation has a long history~\cite{badler1993simulating, brand2000style, wang2007gaussian}. Most approaches generate the motion of 3D skeletons~\cite{gopalakrishnan2019neural, holden2016deep}, while others focus on animating human models like SMPL~\cite{loper2015smpl}. Early work in this field introduced various neural generative models, including autoencoders~\cite{plappert2016kit}, Variational Autoencoders (VAEs)~\cite{petrovich2022temos}, and Generative Adversarial Networks (GANs)~\cite{xu2023actformer}, laying the groundwork for significant advancements in human motion generation. Recently, the advantages of diffusion models~\cite{sohl2015deep} have garnered attention in various domains, such as image generation~\cite{ho2020denoising}, video generation~\cite{yang2023diffusion}, audio synthesis~\cite{kong2020diffwave}, and hand reconstruction~\cite{ye2023diffusion}. In the domain of human motion generation, recent works~\cite{huang2024stablemofusion, tevet2023human, guo2022generating} have employed diffusion models~\cite{sohl2015deep}, achieving impressive results. A variety of enhancements have been proposed to further optimize these models, including incorporating physics constraints~\cite{yuan2023physdiff}, scene awareness~\cite{huang2023diffusion}, improving generation efficiency through latent space diffusion~\cite{chen2023executing}, and refining control conditions via large language models~\cite{zhang2023finemogen}.

Despite these advancements, most methods focus on generating isolated human motions without considering interactions with objects. The InterDiff model~\cite{xu2023interdiff} addresses this limitation by generating human-object interactions using diffusion models. This approach has been further refined in recent works~\cite{peng2023hoi, li2024genzi} that incorporate contact-based predictions along with inference-time guidance to enhance interaction quality. However, these methods concentrate on full-body motions and overlook intricate hand-object interactions. In contrast, our work focuses on detailed hand-object interactions within a full-body pose. The most similar concurrent work~\cite{wu2024human} also leverages a diffusion model for human-object interaction synthesis; however, their approach is limited to interactions with a single object in a fixed pose. Our approach, in comparison, aims for broader generalization, enabling the synthesis of various interaction scenarios conditioned on textual descriptions.

\subsection{Human-Object Interaction Generation}
Recent advances in interaction synthesis have been significantly driven by the availability of high-quality human-object interaction datasets~\cite{brahmbhatt2018contactdb, chao2021dexycb, jian2023affordpose, taheri2020grab, bhatnagar22behave}. Some studies focus on generating coarse full-body object interactions~\cite{luo2021dynamics, peng2023hoi, li2023object}, such as carrying or moving objects. For example, FLEX~\cite{tendulkar2022flex} trains a prior model for hand and body poses and later optimizes these priors to produce a range of static full-body grasps. GOAL~\cite{taheri2021goal} and SAGA~\cite{wu2022saga} employ conditional variational autoencoders (CVAEs) to generate approach motions for full-body grasps, while COOP~\cite{Zheng_2023_ICCV} decouples the full-body pose into separate body and hand poses, which are then coupled through a unified optimization algorithm. Similarly, Braun et al.~\cite{braun2024physically} model full-body motion dynamics using physics simulation and reinforcement learning. 

Another research direction focuses on generating hand-object interaction sequences independently of full-body motion, motivated by applications in virtual reality (VR) and the need for models capable of capturing fine-grained finger motion details. A prevalent approach involves using physical simulation and reinforcement learning~\cite{christen2022d, garcia2020physics, wang2023dexgraspnet}. Some studies address dexterous manipulation by learning from full human demonstration data, either collected via teleoperation~\cite{rajeswaran2017learning} or from video recordings~\cite{garcia2020physics, qin2022dexmv}. For instance, Mandikal et al.~\cite{mandikal2021learning} propose a reward function that encourages robotic hand policies to grasp objects within their affordance regions. D-Grasp~\cite{christen2022d} leverages reinforcement learning and physics simulation to generate diverse hand-object interactions based on sparse reference inputs, while ArtiGrasp~\cite{zhang2024artigrasp} extends this framework to two-handed grasps and generates articulated object motions.

Most existing approaches address only parts of the problem (e.g., focusing on either the body or the hands) or fail to generate diverse results under the same condition settings. In contrast, our work emphasizes fine-grained human-object interactions with the capability to generalize to unseen objects. Furthermore, we leverage language prompts as input to guide interaction generation. Among methods modeling full-body object interactions, the most comparable is TOHO~\cite{li2024task}, which synthesizes approach and manipulation tasks using neural implicit representations. However, TOHO lacks diversity in generated poses under identical conditions and struggles with generating appropriate poses for out-of-domain (OOD) data. Our model overcomes these limitations by refining poses with contact-based guidance and employing dynamic adaptation for OOD scenarios, enabling diverse contact poses under similar conditions and robust generalization to unseen objects.

\begin{figure*}[!tbp]
\centering
\includegraphics[width=\textwidth, trim=0cm 5.5cm 0cm 2.8cm, clip]{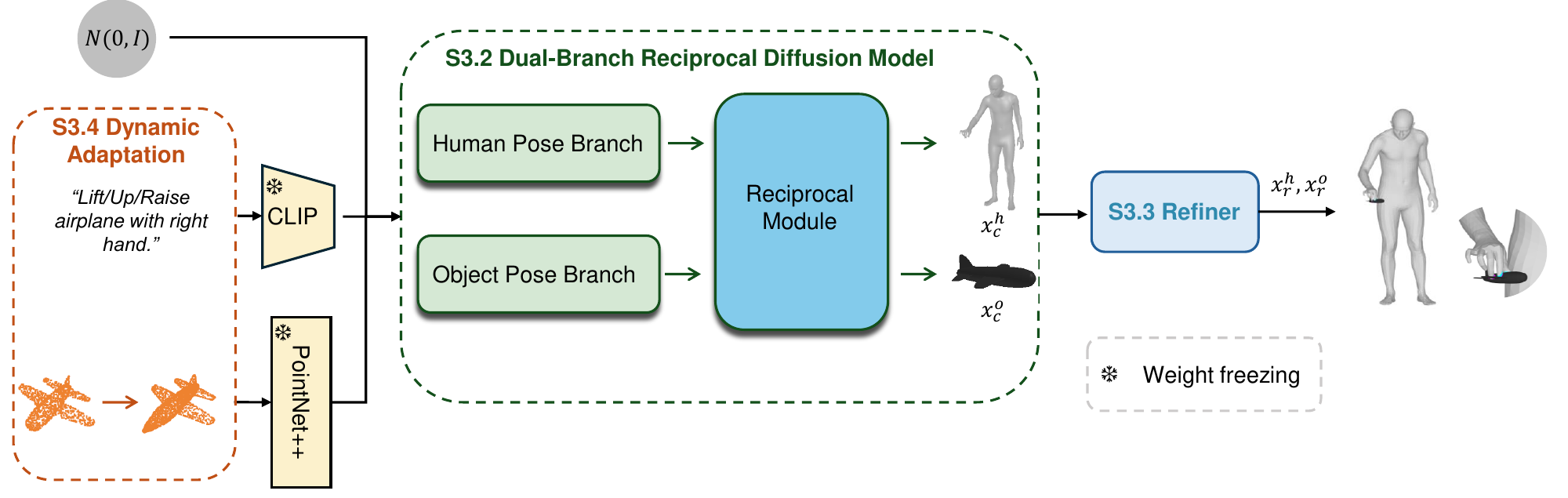}
\caption{Overview of OOD-HOI. Our approach decomposes the generation process into three module: (1) a dual-branch reciprocal diffusion model that exchanges information between human and object to generate an initial interaction pose, (2) a contact-guided interaction refiner is employed to revise the initial interaction human-object pose with additional inference-time guidance, (3) and a dynamic adaptation module designed for out-of-domain (OOD) generation, ensuring more realistic and physically plausible results.}
\label{fig:method}
\vspace{-.5em}
\end{figure*}

\section{Method}
\subsection{Overview}

In this work, we introduce a novel approach for text-driven, whole-body human-object interactions, specifically for out-of-domain (OOD) generation tasks. Our method, named \mymethod{OOD-HOI}, is designed to jointly consider information exchange between the human body, hands, and objects to achieve a cohesive interaction, as shown in Fig.~\ref{fig:method}. Our framework consists of three key components: a dual-branch reciprocal diffusion model, a contact-guided interaction refiner and dynamic adaptation. The dual-branch reciprocal diffusion model, described in Sec.~\ref{sec:coarse generation}, generates compositional whole-body interactions based on text descriptions and object point clouds. Subsequently, the contact-guided interaction refiner, detailed in Sec.~\ref{sec:interaction refiner}, adjusts the interaction pose by utilizing predicted contact areas as guidance. This refiner also allows for additional inference-time guidance through a diffusion process to enhance the generated poses. To improve the model's generalization for previously unseen objects and diverse text descriptions, we incorporate dynamic adaptation in Sec.~\ref{sec:data aug}. It includes semantic adjustment and geometry deformation modules to enable the generalization of our method for more robust and adaptable interaction generation.


\subsection{Dual-Branch Reciprocal Diffusion Model}
\label{sec:coarse generation}
The 3D human-object pose is represented as $x_0 = \{x^h_0, x^o_0\}$, where the overall pose comprises two elements: the human body pose and the object pose. Specifically, the human pose $x^h_0 \in \mathbb{R}^{159}$ is represented by a 159-dimensional vector in SMPLH~\cite{MANO:SIGGRAPHASIA:2017}, while the object pose $x^o_0 \in \mathbb{R}^6$ is a 6-dimensional vector~\cite{taheri2020grab}. Due to the differing distributions between human and object poses, we propose a Transformer-based dual-branch reciprocal model to generate coherent human and object poses. This model decomposes the overall pose generation into two distinct Transformer modules~\cite{vaswani2017attention}: human pose branch and object pose branch. This modular approach enhances the generation process by ensuring the rationality of each component’s posture.





The Reciprocal Module (RM) is structured as a Transformer block, operating on intermediate features $f = \left\{f^h, f^o\right\}$, where $f^h$ and $f^o$ represent the human and object features. Each input feature is obtained from a diffusion encoder with appropriate conditions, and the RM refines these features through a cross-attention mechanism~\cite{chen2021crossvit}. This cross-attention mechanism allows the module to retain essential interaction information by refining $f^h$ and $f^o$ in a mutually conditioned manner. Specifically, the RM updates the input encoded features to generate new representations $\hat{f} = \left\{\hat{f}^h, \hat{f}^o\right\}$, where each updated feature, $\hat{f}^h$ and $\hat{f}^o$, is conditioned on the other. These updated features are then passed to the subsequent layers of their respective branches, ultimately producing the final, refined poses $x = \left\{x^h, x^o\right\}$, where $x^h$ and $x^o$ account for the distinct dimensionalities of human and object poses.
%
%
\begin{align}
\hat{f}^h = \text{Cross-Attention}(Q_{f^h}, K_{f^h}, V_{f^o}) 
\\
\hat{f}^o = \text{Cross-Attention}(Q_{f^o}, K_{f^o}, V_{f^h})
\end{align}

Here, $Q_{f^h}$, $K_{f^h}$, $V_{f^h}$ represent the query, key, and value matrices derived from the human features $f^h$, with similar definitions for the object features $f^o$. The integration of the RM enables a more precise human-object interaction in the generated poses; this design choice is evaluated in Tab.~\ref{tab:Ablation}. Further methodological details of the dual-branch reciprocal diffusion model are provided in the supplementary material.


\subsection{Contact-Guided Interaction Refiner}
\label{sec:interaction refiner}

\begin{figure}[!tbp]
\centering
\includegraphics[width=0.48\textwidth, trim=0cm 5cm 1cm 0cm, clip]{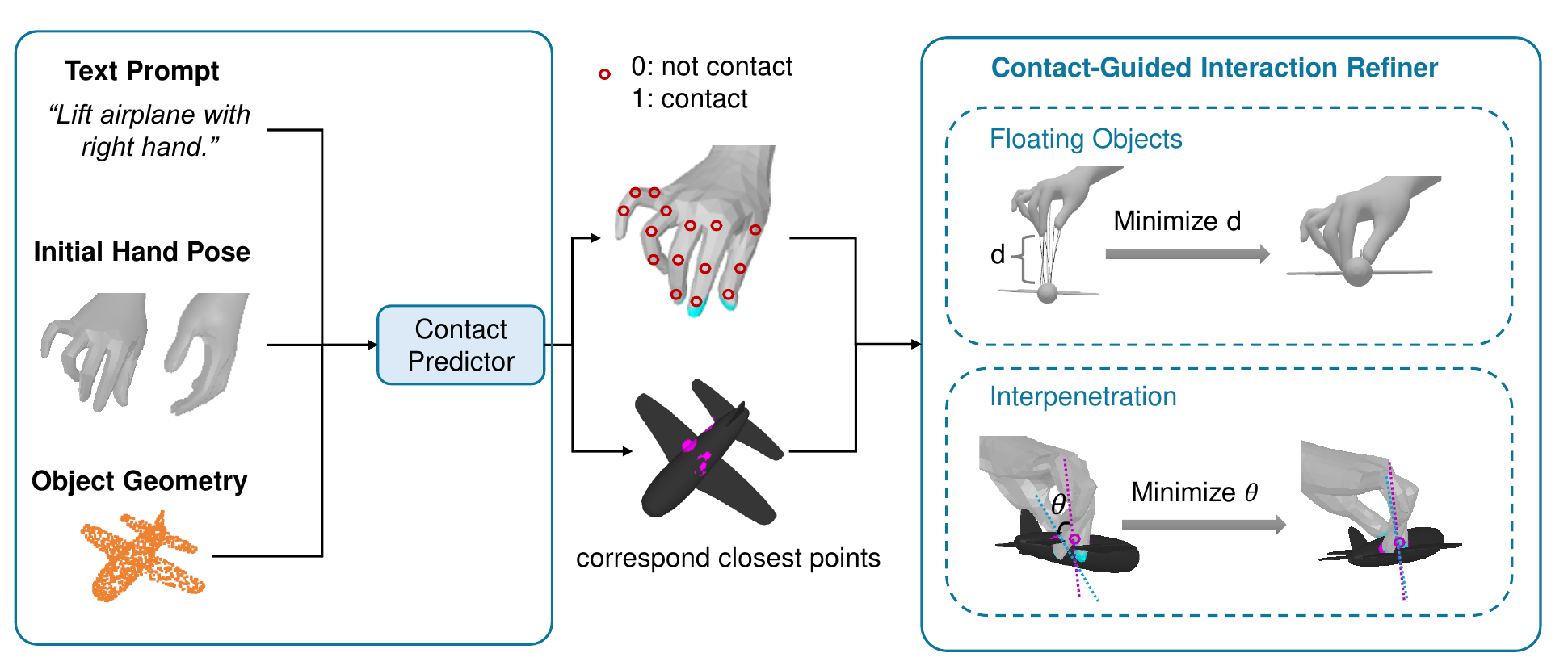}
\caption{Contact-Guided Interaction Refiner to conduct physical optimization. The refiner module takes text prompt, initial hand pose and object geometry as input, predicts the contact area between hand and object, and optimizes the floating object and interpenetration based on the predicted contact areas.}
\label{fig:contact_guided_refiner}
\vspace{-.5em}
\end{figure}

Due to the complexity inherent in human-object interactions, initial generated results often lack physical plausibility~\cite{peng2023hoi, taheri2021goal}. These results frequently exhibit artifacts such as floating objects or interpenetration. To enhance the synthesis of intricate interactions, we propose a contact-guided interaction refiner that addresses interaction issues by incorporating predicted contact areas as guidance.

Different from previous works optimize interactions from downsampled vertex distance~\cite{ghosh2022imos, taheri2021goal,taheri2020grab}, we introduce a guidance function, $F(\mu^h_t, \mu^o_t, y_0)$, which evaluates the alignment between the 30 hand joints and the object's 6 DoF pose. Here, $\mu^h_t \in \mathbb{R}^{30 \times 3}$ is hand joints 
, $\mu^o_t \in \mathbb{R}^{4000 \times 3}$ represents object points, and $y_0 \in \mathbb{R}^{30 \times 4}$ denotes the predicted contact area along with the contact probability. Our guidance function $F(\mu^h_t, \mu^o_t, y_0)$ consists of two main components that are used to optimize the interaction:
\begin{equation}
\label{eq_full}
F(\mu^h_t, \mu^o_t, y_0) = F_{\text{con}} + 0.1 \times F_{\text{norm}}
\end{equation}
Here, $F_{con}$ is contact distance function aiming to reducing object floating and $F_{norm}$ is normal vector function focusing on interpenetration. The detailed definitions of these terms are as follows.

To ensure proximity between human and object contact points, we minimize the distance between the hand contact vertices and the object contact points. Given the predicted contact area $y_0$, we use fixed mapping points for the hand joints, since hand deformation is relatively minor compared to body shape deformation. This objective is formulated as follows:
\begin{equation}
\label{eq_con}
F_{\text{con}} = \sum_{i=1}^{30} \left| R(\mu^h_t(i)) - V(y^o_t(i)) \right|^2,
\end{equation}
where $\mu^h_t(i)$ and $y^o_t(i)$ represent the $i$-th contacting vertex of the hand and the corresponding object contact point, respectively. The function $R(\cdot)$ transforms the human vertices to object coordinates, while $V(\cdot)$ retrieves the object contact points based on the predicted contact area $y^o_t(i)$.

While the contact proximity term reduces floating artifacts, it cannot fully prevent interpenetration. Interpenetration often arises from incorrect contact angles. To mitigate this issue, we introduce a normal vector alignment criterion at the contact points. Specifically, we enforce alignment by requiring that the sum of the normal vectors at the contact points is zero, ensuring contact remains on the object's surface. This objective is formulated as:
\begin{equation}
\label{eq_norm}
F_{\text{norm}} = \sum_{i=1}^{30} \left| N_h R(\mu^h_t(i)) + N_o V(y^o_t(i)) \right|^2,
\end{equation}
where $N_h$ and $N_o$ represent the normal vectors of corresponding contact human vertices and object points.

By optimizing Eq.~\ref{eq_full}, our approach effectively addresses both floating and penetration artifacts, leading to realistic and physically plausible human-object interactions.

\subsection{Dynamic Adaptation}
\label{sec:data aug}
Given that the human-object interactions dataset is significantly smaller than other large-scale datasets (e.g., ImageNet~\cite{deng2009imagenet}) and the diversity of interaction types in daily life is vast, it is essential to address the out-of-domain (OOD) generalization problem. The OOD generalization challenge arises when the trained model encounters inputs that deviate substantially from its training data, such as novel objects, unconventional actions, or previously unseen human-object combinations. In the context of human-object interactions, the OOD generation problem typically manifests in two key aspects: Text-OOD and Object-OOD. To address these aspects of OOD generation problem, we propose a dynamic adaptation strategy, which incorporates semantic adjustment for Text-OOD and geometry deformation for Object-OOD.

\textbf{Semantic Adjustment.} When the model encounters descriptions of unconventional actions, it often struggles to generate appropriate human motions. To address this, we propose a semantic adjustment mechanism that enhances the diversity of input text semantics. Specifically, we adjust action intent by replacing verbs with synonyms using GPT-4o (e.g., substituting \textit{lift} with synonyms like \textit{raise}, \textit{up}, or \textit{uplift}). During training, we randomly combine these synonym variations with object and hand-contact descriptions. This process improves the model’s ability to interpret and respond to unconventional actions by expanding its semantic range and adaptability.

\begin{figure}[!tbp]
\includegraphics[width=0.8\textwidth, trim=20 70 0 20, clip]{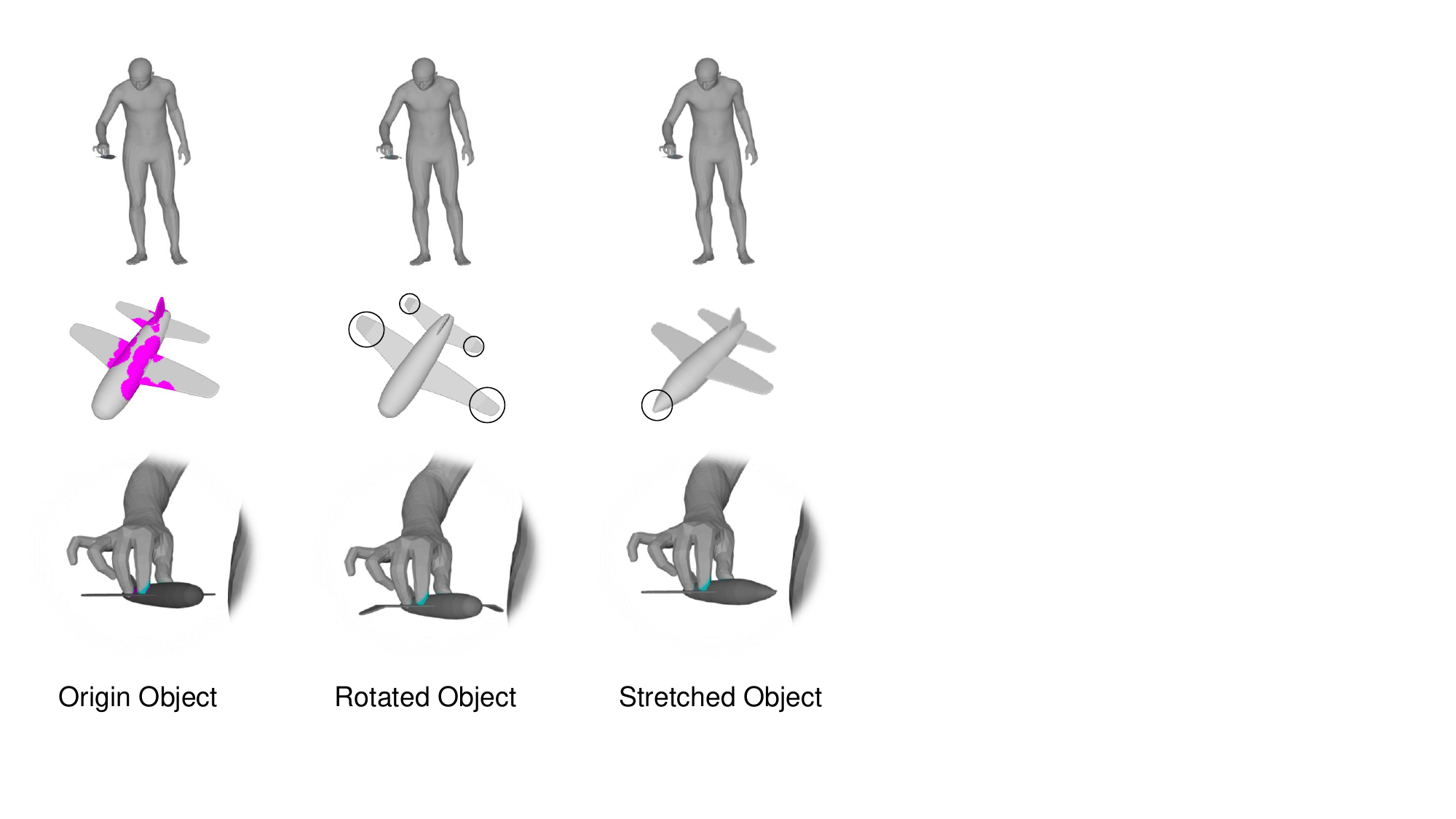}
\caption{For geometry deformation, we propose a condition enhancement that deforms the object under a constraint of constant contact area. Since the primary contact for the airplane model typically occurs on its body, we apply controlled random deformations such as rotating the wings or stretching the nose within specified limits, which improve model robustness.}
\label{fig:deformation}
\vspace{-.5em}
\end{figure}

\begin{table*}[!tbp]
\caption{Result of Quantitative Evaluation. We compare our methods with five state-of-art methods on GRAB dataset~\cite{taheri2020grab}. The best results are emphasized in \textbf{bold}.}
\centering
\label{tab:Quantitative Evaluation}
\tabcolsep=10pt
\begin{tabular}{lccccc}
\toprule
Method & Accuracy (top-3) $\uparrow$ & FID $\downarrow$ & Diversity $\uparrow$ & Multimodality $\uparrow$ & IV$[cm^3]$ $\downarrow$\\ 
\midrule
GT & 0.999 & - & 0.856 & 0.439 & 3.40 \\ 
\midrule
T2M~\cite{guo2022generating} $_{CVPR 2022}$ & 0.190 & 0.789 & 0.571 & 0.096 & -\\ 
MDM~\cite{tevet2023human} $_{ICLR 2023}$ & 0.515 & 0.625 & 0.441 & 0.339 & 16.34\\ 
IMoS~\cite{ghosh2022imos} $_{EuroGraph 2022}$  & 0.410 & 0.615 & 0.686 & 0.284 & 7.14 \\ 
DiffH2O~\cite{christen2024diffh2o} $_{SigAsia 2024}$ & 0.833 & 0.834 & 1.144 & 0.318 & 6.02\\ 
Text2HOI~\cite{cha2024text2hoi} $_{CVPR 2024}$ & 0.922 & 0.301 & \textbf{0.835} & 0.522 & 12.43 \\ 
\midrule
\textbf{Ours} & \textbf{0.933} & \textbf{0.213} & 0.823 & \textbf{0.540} & \textbf{3.15}\\  
\bottomrule
\end{tabular}%
\end{table*}

\textbf{Geometry Deformation.} Interaction with novel objects requires the model to identify plausible contact areas for realistic human-object engagement. Without appropriate adaptation, the model may inaccurately predict contact points, leading to unrealistic interactions. To address this, we apply geometry deformation. We define the usual contact region of an object as the pre-labeled contact area associated with typical interactions described in the text (e.g., grasping a mug typically involves the middle of its handle, as opposed to grasping both sides of the handle). Our approach deforms parts of the object outside this usual contact area with a specified probability and degree of deformation. The augmented object, denoted as $p_a$, is formulated as follows:
\begin{equation}
    p_a = \eta \times (p_{uno} \times d_f) + (p_o - p_{uno})
\end{equation}
where $p_o$ represents the original object point clouds, $\eta$ is the probability of deformation being applied, $p_{uno}$ indicates the less frequently used contact regions of the object, and $d_f$ is the degree of deformation. As illustrated in Fig.~\ref{fig:deformation}, the typical contact area, shown in purple, represents the airplane body. To generate additional training data, we introduce controlled deformations to a typical contact areas, such as by rotating the wings or elongating the nose. This process enables the model to learn from a broader range of human-object interactions, thereby enhancing its robustness in generating realistic interactions across varying levels of deformation.

Together, these two approaches allow the model to adjust its understanding of unconventional actions and adapt to the geometric characteristics of previously unseen objects, improving both its generalization and adaptability in out-of-domain scenarios.

\begin{table}[!tbp]
\caption{Result of out-of-domain Quantitaive Evaluation. We compare our methods with three state-of-art methods on HO-3D dataset~\cite{hampali2020honnotate}. Best results are emphasized in \textbf{bold}.}
\centering
\label{tab:OOD Quantitative Evaluation}
\resizebox{0.48\textwidth}{!}{%
\tabcolsep=1pt
\begin{tabular}{lccccc}
\toprule
Method & Acc. (top-3) $\uparrow$ & Diversity $\uparrow$ & Multimodality $\uparrow$ & IV$[cm^3]$ $\downarrow$\\ 
\midrule
IMoS~\cite{ghosh2022imos}  & 0.581 & 1.05 & 0.22 & 10.38 \\ 
DiffH2O~\cite{christen2024diffh2o} & 0.755 & \textbf{1.09} & 0.23 & 9.03\\ 
Text2HOI~\cite{cha2024text2hoi} & 0.735 & 0.44 & 0.23 & 11.02 \\ 
\midrule
\textbf{Ours (w/o DA)} & 0.750 & 0.45 & 0.22 & 7.49 \\  
\textbf{Ours} & \textbf{0.852} & 0.46 & \textbf{0.33} & \textbf{6.41} \\  
\bottomrule
\end{tabular}%
}
\end{table}

\section{Experiments}
\subsection{Datasets}
We utilize the subject-based split of the GRAB dataset~\cite{taheri2020grab}, as proposed in IMoS~\cite{cha2024text2hoi}, to facilitate direct comparison. Additionally, we test semantic generalization by introducing unseen synonyms actions. However, since this split does not include any unseen objects, we incorporate objects from the HO-3D dataset~\cite{hampali2020honnotate} to further evaluate guidance effectiveness and object-level generalization. More details of dataset are provided in the supplementary material.

\subsection{Evaluation Results}

\subsubsection{Evaluation Metrics and Baselines}
\textbf{Evaluation metrics.} Following previous work on human motion synthesis~\cite{cha2024text2hoi,tevet2023human}, we use the metrics of accuracy, frechet inception distance (FID), diversity and multi-modality, as used in Text2HOI~\cite{cha2024text2hoi}. The accuracy serves as an indicator of how well the model generates poses with text and is evaluated by the pre-trained action classifier. We train a standard RNN-based action classifier to extract pose features and classify the action from the poses, as in IMoS~\cite{ghosh2022imos}. The FID quantifies feature-space distances between real and generated poses, capturing the dissimilarity. The diversity reflects the range of distinct poses, and multi-modality measures the average variance of poses for an individual text prompt. Moreover, we use intersect volume to evaluate the interaction between human and object, as used in DiffH2O~\cite{christen2024diffh2o}, which indicated the number of hand vertices that penetrate the object mesh.


\begin{figure}[!tbp]
\vspace{-1.2em}
\centering
\includegraphics[width=0.5\textwidth, trim=30 60 250 20 , clip]{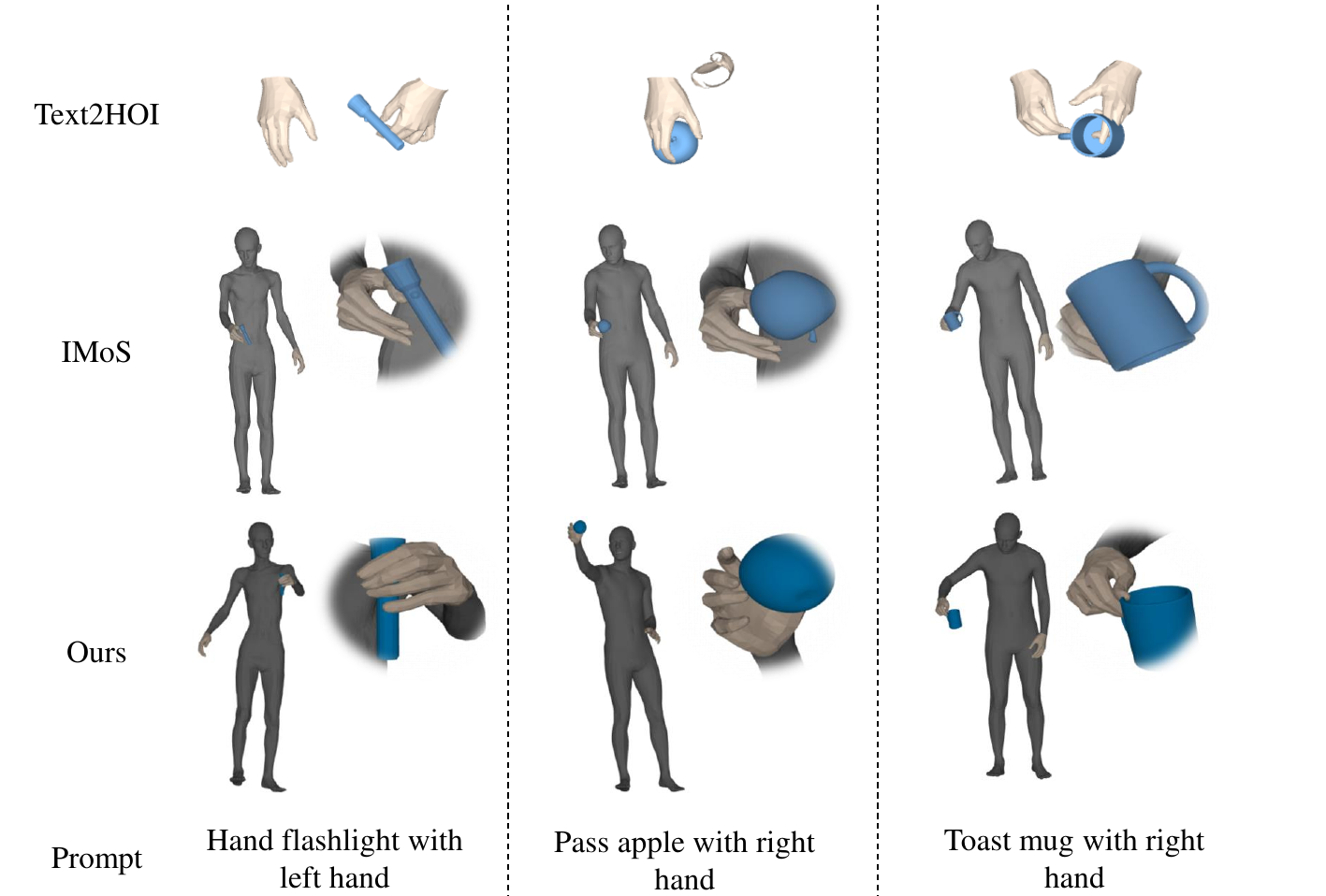}
\caption{We compare our generated human-object interaction pose with other baseline results in GRAB dataset~\cite{taheri2020grab}. Each row show the results of Text2HOI~\cite{cha2024text2hoi}, IMoS~\cite{ghosh2022imos}, and Ours.}
\label{fig:in_result}
\vspace{-1.5em}
\end{figure}

\begin{figure*}[!tbp]
\centering
\includegraphics[width=1.12\textwidth, trim=10 60 30 0 , clip]{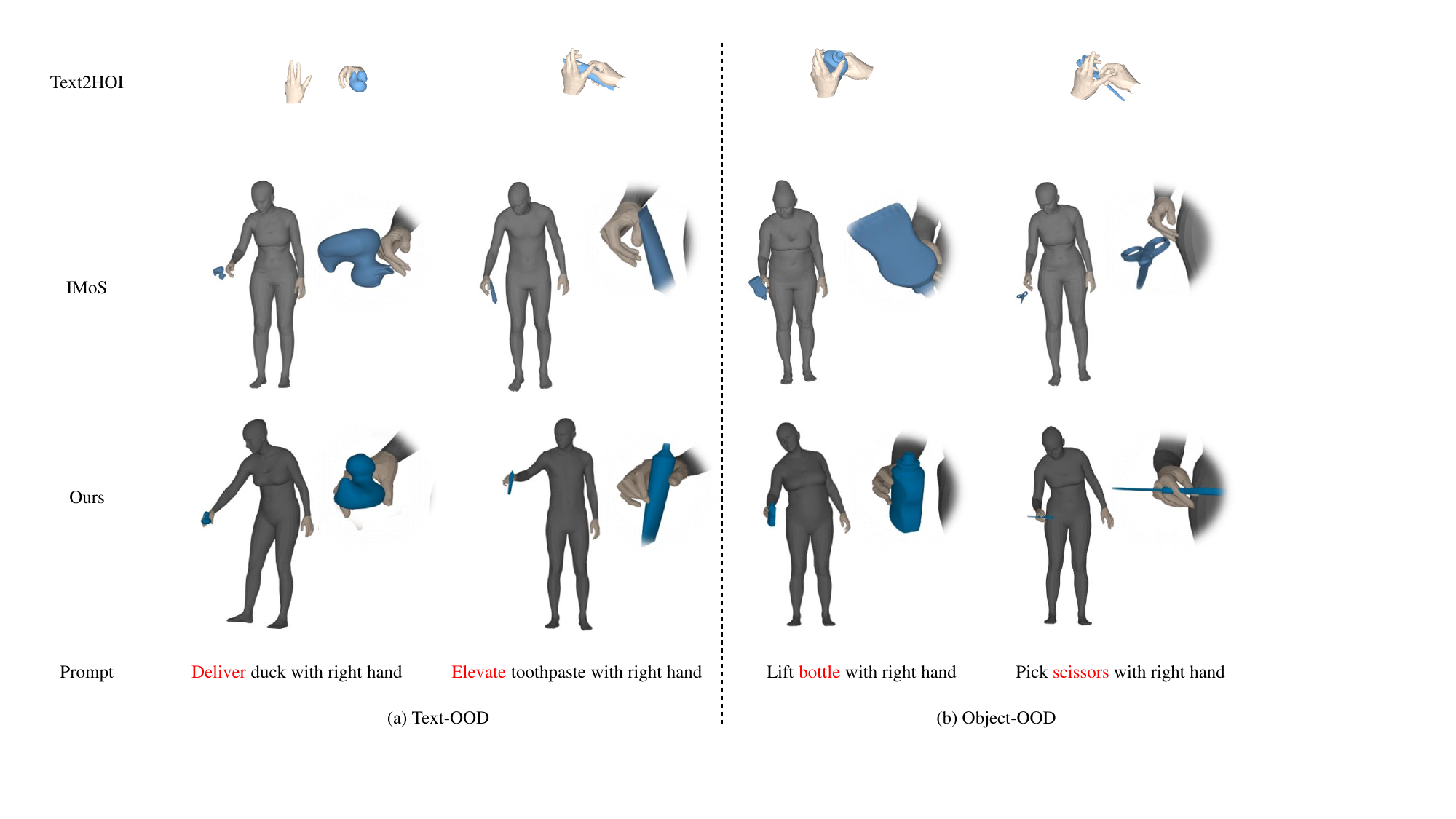}
\caption{Qualitative Comparison of out-of-domain (OOD) Performance. We validate our method across two OOD scenarios (highlighted in red): (a) Text-OOD and (b) Object-OOD, against other baseline methods. Results demonstrate that our method not only accurately generates contact points but also effectively conveys the intended action.}
\label{fig:ood_result}
\vspace{-.6em}
\end{figure*}

\textbf{Baselines.} We compare our approach with five existing text-to-human motion generation methods: T2M~\cite{guo2022generating}, MDM~\cite{tevet2023human}, IMoS~\cite{ghosh2022imos}, DiffH2O~\cite{christen2024diffh2o} and Text2HOI~\cite{cha2024text2hoi}. T2M~\cite{guo2022generating} employs a temporal VAE-based architecture and MDM~\cite{tevet2023human} utilizes a transformer based diffusion model. IMoS~\cite{ghosh2022imos} is designed to generate human body and arm motions based on both action labels and past body motions. DiffH2O~\cite{christen2024diffh2o} first utilizes a UNet-based diffusion model to synthesize text and hand-object interactions. Moreover, Text2HOI generates sequences of hand-object interaction within the object contact condition.

\subsubsection{Quantitative Evaluation}

\textbf{In-Domain Analysis.} Tab.~\ref{tab:Quantitative Evaluation} presents quantitative metrics, with results for IMoS~\cite{ghosh2022imos} and DiffH2O~\cite{christen2024diffh2o} sourced from DiffH2O~\cite{christen2024diffh2o}, and T2M~\cite{guo2022generating} and Text2HOI~\cite{cha2024text2hoi} from Text2HOI~\cite{cha2024text2hoi}. Our method achieves the lowest FID scores, indicating high-quality pose generation that closely matches ground truth actions. As emphasized in~\cite{huang2024stablemofusion}, diversity and multimodality are relevant only when poses are realistic; hence, among models with low FID (Text2HOI and Ours), ours also excels in diversity and one-to-many mappings. For the IV metric, our model outperforms Text2HOI, effectively managing interpenetration issues. Additionally, our model ranks highest in accuracy, aligning generated poses closely with textual descriptions. Overall, our method leads in action quality, diversity, physical realism, and text alignment.

\textbf{Out-of-Domain Analysis.} The quantitative metrics for out-of-domain data are shown in Tab.~\ref{tab:OOD Quantitative Evaluation}. As with in-domain metrics, the experimental results of IMoS~\cite{ghosh2022imos} and DiffH2O~\cite{christen2024diffh2o} are sourced from DiffH2O~\cite{christen2024diffh2o}. To evaluate model performance on out-of-domain data, we conduct quantitative experiments on the HO-3D~\cite{hampali2020honnotate} dataset. Our method achieves the best results across all metrics, indicating a significant improvement over other approaches on out-of-domain data. This improvement underscores the effectiveness of our proposed dynamic adaptation module in enhancing the model's generalization capabilities.

\begin{table*}[!tbp]
\caption{Ablation study on model structure, refiner function, and dynamic adaptation. We compare our model with four configurations: single-branch (Single), dual-branch without reciprocal (Dual), dual-branch with encoder-based reciprocal (RM-enc), and our dual-branch with decoder-based reciprocal (RM). We also show results for each refiner function applied individually and analyze the two aspects of dynamic adaptation. The best results are in \textbf{bold}.}
\centering
\label{tab:Ablation}
\tabcolsep=13pt
\begin{tabular}{lccccc}
\toprule
Method & Accuracy (top-3) $\uparrow$ & FID $\downarrow$ & Diversity $\uparrow$ & Multimodality $\uparrow$ & IV$[cm^3]$ $\downarrow$\\ 
\midrule
Single & 0.515 & 0.625 & 0.441 & 0.339 & 16.34 \\
Dual & 0.617 & 0.483 & 0.483 & 0.269 & 14.88 \\
RM-enc & 0.685 & 0.553 & 0.562 & 0.295 & 13.23 \\
RM & 0.883 & 0.383 & 0.465 & 0.372 & 11.02 \\ 
\midrule
w/o $F_{con}$ & 0.884 & 0.380 & 0.462 & 0.374 & 8.33 \\ 
w/o $F_{norm}$ & 0.882 & 0.385 & 0.474 & 0.370 & 4.69 \\ 
\midrule
w/o GD & 0.911 & 0.337 & 0.715 & 0.474 & 3.42\\ 
w/o SA & 0.924 & 0.305 & 0.746 & 0.493 & 3.22\\ 
\midrule
\textbf{w GD$\&$SA} & \textbf{0.933} & \textbf{0.213} & \textbf{0.823} & \textbf{0.540} & \textbf{3.15}\\  
\bottomrule
\end{tabular}%
\vspace{-.5em}
\end{table*}

\subsubsection{Perceptual Evaluation}
To evaluate the visual quality of our interaction pose, we conduct a perceptual study where we compare our results with two state-of-the-art methods Text2HOI~\cite{cha2024text2hoi} and IMoS~\cite{ghosh2022imos}. Text2HOI only generates hand pose from text while IMoS generates full body contact through action intent. We conduct our perceptual study in the following two sections.

\textbf{In-Domain Analysis.} As shown in Fig.~\ref{fig:in_result}, our method significantly outperforms the other two approaches. For instance, given the input text \textit{toast mug with right hand}, only our approach successfully grasps the mug with a realistic pose, proper contact points, and the correct motion intent. In contrast, Text2HOI exhibits severe interpenetration with the left hand, while IMoS fails to make contact with the object’s surface. These results demonstrate that, unlike previous methods, our approach uniquely generates physically plausible and realistic interaction poses.

\textbf{Out-of-Domain Analysis.} 
As shown in Fig.~\ref{fig:ood_result}, we present results for text-based out-of-domain (Text-OOD) and object-based out-of-domain (Object-OOD) tests, demonstrating our method's strengths over existing alternatives. For the Text-OOD test, we assess our model's handling of unseen terms \textit{deliver} and \textit{elevate}. Our approach successfully interprets and generates distinct pose for these terms, while Text2HOI struggles with \textit{elevate}, producing less accurate responses, and IMoS generates nearly identical poses for both terms, showing limited understanding of intent. In the Object-OOD test, we use the \textit{bottle} and \textit{scissors} from the HO-3D dataset~\cite{hampali2020honnotate} to evaluate object generalization. Our model performs well, accurately reflecting the physical properties of each object. In contrast, Text2HOI faces interpenetration issues, especially in non-contact hand handling, and IMoS fails to establish proper contact with objects. These results highlight our method’s robustness in both Text-OOD and Object-OOD scenarios, consistently outperforming competing methods with accurate, context-sensitive motions and controlled object interactions.

\subsection{Ablation}

We conduct several ablation studies on GRAB dataset~\cite{taheri2020grab}, to valid the effectiveness of out modules. The results are demonstrated in Tab.~\ref{tab:Ablation}.

\textbf{Model Structure.} We evaluate our model structure from two different perspectives. First, to examine the advantages of our dual-branch structure, which decouples human and object poses, we compare it with a single-branch MDM model~\cite{tevet2023human} (denoted as ``Single''). The dual-branch structure provides more dedicated feature representations for human and object poses, resulting in enhanced realism and physical plausibility in the generated interactions. Next, we assess the effectiveness of the reciprocal model (denoted as ``RM'' and ``Dual'') by modifying it to replace the decoder structure with an encoder (referred to as ``RM-enc''). This comparison highlights the contribution of cross-attention in refining interaction quality.

\textbf{Refiner Function.} We analyze the effects of removing the contact distance function, ``$F_{con}$'', and normal vector function, ``$F_{norm}$'', from our refiner by evaluating the model variants ``w/o $F_{norm}$'' and ``w/o $F_{con}$''. From these ablation studies, we conclude that both the contact distance function and normal vector function are crucial for improving the model’s understanding of the 3D spatial relationships between the human and objects, leading to more realistic and physically consistent interactions.

\textbf{Dynamic Adaptation.} We examine the impact of geometry deformation (GD) and semantic adjustment (SA) by comparing our full model (denoted as ``w GD$\&$SA'') with a variant without geometry deformation (``w/o GD'') and another without semantic adjustment (``w/o SA''). Our model, which includes both geometry deformation and semantic adjustment, demonstrates superior performance in terms of multimodality and reduced interpenetration volume. Additionally, these dynamic adaptation improves generalization to out-of-domain datasets, as observed on the HO-3D dataset~\cite{hampali2020honnotate} (see Tab.~\ref{tab:OOD Quantitative Evaluation}).
\section{Conclusion}
In this paper, we propose a novel text-driven method for generating whole-body human-object interactions by jointly considering information exchange among the human body, hands, and objects. This is achieved through a two-stage framework: (1) generating an initial interaction pose using a dual-branch reciprocal diffusion model, and (2) refining the initial human-object pose based on the predicted contact areas. To address the out-of-domain problem, we introduce dynamic adaptation, which includes semantic adjustment and geometry deformation. Experimental results validate the effectiveness of our method on both in-domain and out-of-domain datasets, demonstrating that our approach outperforms state-of-the-art baselines with enhanced physical plausibility and robustness.

\clearpage
{
    \small
    \bibliographystyle{ieeenat_fullname}
    \bibliography{main}
}

\end{document}